%% file: paper.tex
\documentclass[10pt,twocolumn,letterpaper]{article}

\usepackage{etex}

\usepackage{pifont}

\usepackage{formattings/hadianeh}
\usepackage{formattings/macros}

\usepackage{tabularx, booktabs}
\usepackage{makecell}

\usepackage{formattings/iccv}

\usepackage{times}
\usepackage{epsfig}
\usepackage{graphicx}
\usepackage{amsmath}
\usepackage{amssymb}

\usepackage{algorithmic}
\usepackage{algorithm}

\usepackage{caption}
\usepackage[labelfont=bf]{caption}
\usepackage{subcaption}
\usepackage[font={sf,footnotesize,bf}]{caption}
\usepackage{booktabs}

 \iccvfinalcopy 

\def\iccvPaperID{6541\vspace{-4ex}} 

\ificcvfinal\fi

\usepackage{fancyhdr}
\usepackage[subtle]{savetrees}

\begin{document}

\title{ReLeQ: A Reinforcement Learning Approach for Deep Quantization of Neural Networks\vspace{-2ex}}

\author{Ahmed T. Elthakeb \quad Prannoy Pilligundla \quad  Fatemehsadat Mireshghallah \quad  Amir Yazdanbakhsh$^*$\\  Hadi Esmaeilzadeh \\
\textbf{A}lternative \textbf{C}omputing \textbf{T}echnologies ({\color[HTML]{0B6121}{\textbf{ACT}}}) Lab\\
University of California San Diego \quad \quad\quad $^*$Google Brain \\
{\sffamily\small {\{\href{mailto:a1yousse@eng.ucsd.edu}{a1yousse},\ \href{mailto:ppilligu@eng.ucsd.edu}{ppilligu}, 
\href{mailto:fmireshg@eng.ucsd.edu}{fmireshg}\}@eng.ucsd.edu} \quad \quad {\href{mailto:ayazdan@google.com}{ayazdan@google.com}} \quad \quad {\href{mailto:hadi@eng.ucsd.edu}{hadi}@eng.ucsd.edu}}} 

{ 
}

\maketitle
\thispagestyle{fancy}

\input{body/abstract}
\input{body/intro}
\input{body/methodology}
\input{body/learning_proc.tex}
\input{body/experimental_setup_new}

\input{body/evaluation_new}

\input{body/related_new}
\input{body/conclusion}
\section*{Acknowledgement}
This work was in part supported by Semiconductor Research Corporation contract \#2019-SD-2884, NSF awards CNS\#1703812, ECCS\#1609823, Air Force Office of Scientific Research (AFOSR) Young Investigator Program (YIP) award \#FA9550-17-1-0274, and gifts from Google, Microsoft, Xilinx, Qualcomm.

{\small
\bibliographystyle{reference/ieee}
\bibliography{paper}
}

\end{document}

%% file: body/abstract.tex
\begin{abstract}
Deep Neural Networks (DNNs) typically require massive amount of computation resource in inference tasks for  computer vision applications. 
Quantization can significantly reduce DNN computation and storage by  decreasing the bitwidth of  network encodings.
Recent research affirms that carefully selecting the quantization levels for each layer can preserve the accuracy while pushing the bitwidth below eight bits.
However, without arduous manual effort, this deep quantization can lead to significant accuracy loss, leaving it in a position of questionable utility.
As such, deep quantization opens a large hyper-parameter space (bitwidth of the layers), the exploration of which is a major challenge.
We propose a systematic approach to tackle this problem, by automating the process of discovering the quantization levels through an end-to-end deep reinforcement learning framework (\releq). 
We adapt policy optimization methods to the problem of quantization, and focus on finding the best design decisions in choosing the state and action spaces, network architecture and training framework, as well as the tuning of various hyperparamters.
We show how \releq can balance speed and quality, and provide an asymmetric general solution for quantization of a large variety of deep networks (AlexNet, CIFAR-10, LeNet, MobileNet-V1, ResNet-20, SVHN, and VGG-11) that virtually preserves the accuracy ($\leq$ 0.3\% loss) while minimizing the computation and storage cost.
%
%
%
%
%
With these DNNs, \releq enables conventional hardware to achieve 2.2$\times$ speedup over 8-bit execution. 
Similarly, a custom DNN accelerator achieves 2.0$\times$ speedup and energy reduction compared to 8-bit runs.
These encouraging results mark \releq as the initial step towards automating the deep quantization of neural networks.
\end{abstract}
\vspace{-3ex}

%% file: body/intro.tex
\section{Introduction}\label{sec:intro}
\vspace{-1ex}

 Deep Neural Networks (DNNs) have made waves across a variety of domains, from image recognition \cite{Krizhevsky2012ImageNetCW} and synthesis, object detection~\cite{objd, objd2}, natural language processing~\cite{NLP}, medical imaging, self-driving cars, video surveillance, and personal assistance~\cite{sirius, bpzip, Hinton, Lee}.
DNN compute efficiency has become a major constraint in unlocking further applications and capabilities, as these models require rather massive amounts of computation even for a single inquiry.
%
%
One approach to reduce the intensity of the DNN computation is to reduce the complexity of each operation.
To this end, quantization of neural networks provides a path forward as it reduces the bitwidth of the operations as well as the data footprint~\cite{Hubara2017QNN, bitfusion, Judd2016StripesBD}.
Albeit alluring, quantization can lead to significant accuracy loss if not employed with diligence.
Years of research and development has yielded current levels of accuracy, which is the driving force behind the wide applicability of DNNs nowadays.
To prudently preserve this valuable feature of DNNs, accuracy, while benefiting from quantization the following two fundamental problems need to be addressed.
(1) learning techniques need to be developed that can train or tune quantized neural networks given a level of quantization for each layer.
(2) Algorithms need to be designed that can discover the appropriate level of quantization for each layer while considering the accuracy.  
This paper takes on the second challenge as there are inspiring efforts that have developed techniques for quantized training~\cite{Zhou2016DoReFaNetTL, Zhu2016TrainedTQ, Mishra2017WRPNWR}.

This paper builds on the algorithmic insight that the bitwidth of operations in DNNs can be reduced \emph{below eight bits} without compromising their classification accuracy.
However, this possibility is manually laborious~\cite{Micikevicius2017MixedPT, Mishra2017ApprenticeUK,DBLP:journals/corr/abs-1812-00090} as to preserve accuracy, the bitwidth varies across individual layers and different DNNs~\cite{Zhou2016DoReFaNetTL, Zhu2016TrainedTQ, Li2016TernaryWN, Mishra2017WRPNWR}.
Each layer has a different role and unique properties in terms of weight distribution.
Thus, intuitively, different layers display different sensitivity towards quantization. 
Over-quantizing a more sensitive layer can result in stringent restrictions on subsequent layers to compensate and maintain accuracy. 
Nonetheless, considering layer-wise quantization opens a rather exponentially large hyper-parameter space, specially when quantization below eight bits is considered.
For example, ResNet-20 exposes a hyper-parameter space of size $8^l=8^{20}>10^{18}$, where $l=20$ is the number of layers and $8$ is the possible quantization levels.
This exponentially large hyper-parameter space grows with the number of the layers making it impractical to exhaustively assess and determine the quantization level for each layer.

To that end, this paper sets out to automate effectively navigating this hyper-parameter space using Reinforcement Learning (RL).
We develop an end-to-end framework, dubbed \releq, that exploits the sample efficiency of the Proximal Policy Optimization~\cite{ppo} to explore the quantization hyper-parameter space.
The RL agent starts from a full-precision previously trained model and learns the sensitivity of final classification accuracy with respect to the quantization level of each layer, determining its bitwidth while keeping classification accuracy virtually intact.
Observing that the quantization bitwidth for a given layer affects the accuracy of subsequent layers, our framework implements an LSTM-based RL framework which enables selecting quantization levels with the context of previous layers' bitwidths.
Rigorous evaluations with a variety of networks (AlexNet, CIFAR, LeNet, SVHN, VGG-11, ResNet-20, and MobileNet) shows that \releq can effectively find heterogenous deep quantization levels that virtually preserve the accuracy ($\leq$0.3\% loss) while minimizing the computation and storage cost.
The results (Table~\ref{table:main_results}) show that there is a high variance in quantization levels across the layers of these networks.
For instance, \releq finds quantization levels that average to 6.43 bits for MobileNet, and to 2.81 bits for ResNet-20. 
%
%
%
With the seven benchmark DNNs, \releq enables conventional hardware~\cite{TVM} to achieve 2.2$\times$ speedup over 8-bit execution. 
Similarly, a custom DNN accelerator~\cite{stripes} achieves 2.0$\times$ speedup and 2.0$\times$ energy reduction compared to 8-bit runs.
These results suggest that \releq takes an effective first step towards automating the deep quantization of neural networks.

%% file: body/methodology.tex
\section{RL for Deep Quantization of DNNs}
\label{sec:rl_quantization}
\vspace{-1ex}
\begin{figure}
 	\centering 	
	\includegraphics[width=0.35\textwidth]{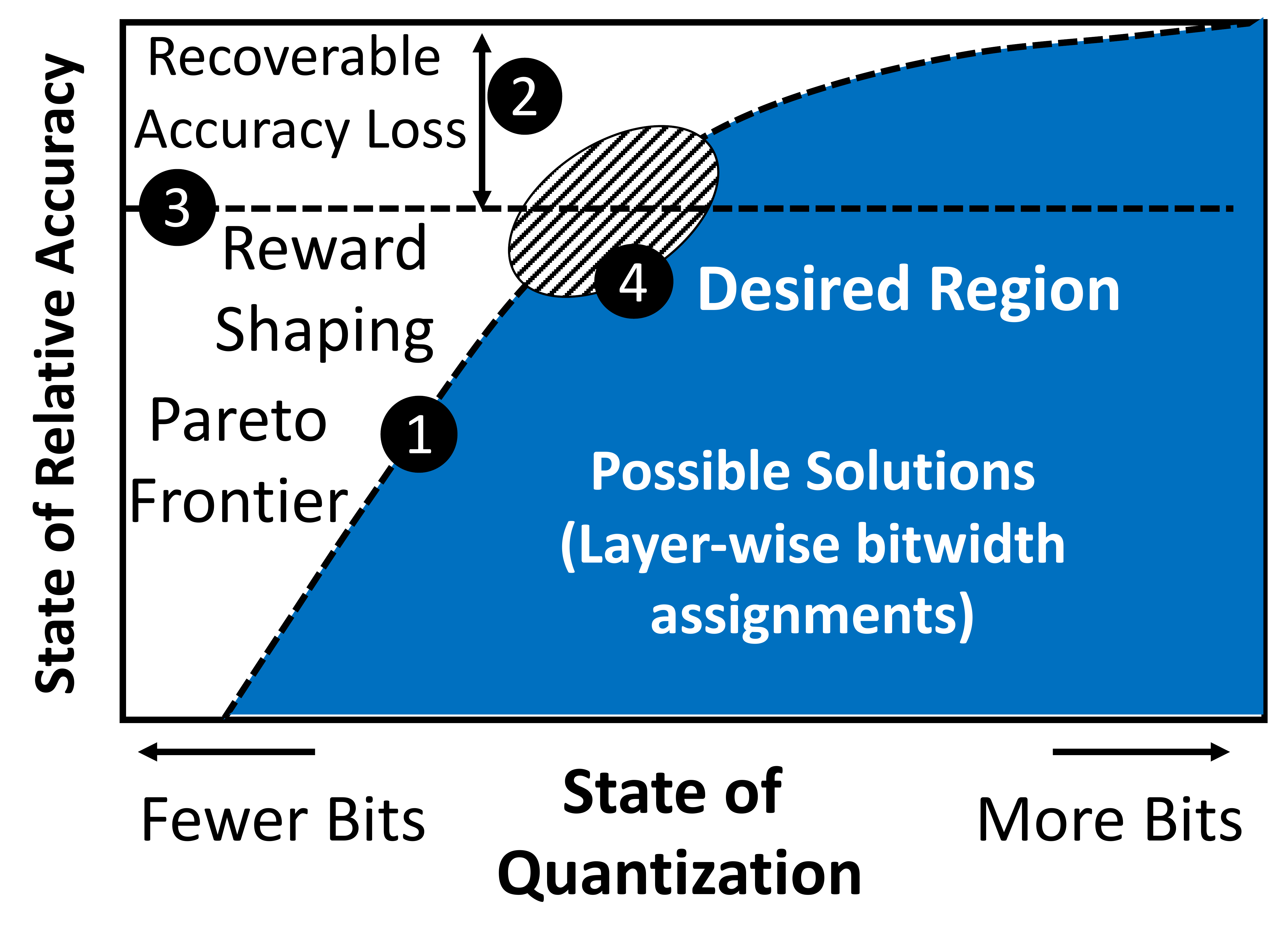}
	\caption{Optimum Automatic Quantization of a Neural Network}
	\label{fig:intro}
 \end{figure}

\subsection{Need for Heterogeneity}
Deep neural networks, by construction, and the underlying training algorithms cast different properties on different layers as they learn different levels of features representations. First, it is widely known that neural networks are heavily overparameterized~\cite{NIPS2019_8847}; thus, different layers exhibit different levels of redundancy. 
Second, for a given initialization and upon training, each layer exhibits a distribution of weights (typically bell-shaped) each of which has a different dynamic range leading to different degrees of robustness to quantization error, hence, different/heterogenous precision requirements. 
Third, our experiments (see Figure~\ref{fig:enum}), where the design space of deep quantization is enumerated for small and moderate size networks, empirically shows that indeed Pareto optimal frontier is mostly composed of heterogenous bitwidths assignment. 
Furthermore, recent works empirically studied the layer-wise functional structure of overparameterized deep models and provided evidence for the heterogeneous characteristic of  layers. 
Recent experimental work~\cite{DBLP:journals/corr/abs-1902-01996} also shows that layers can be categorized as either ``ambient'' or ``critical'' towards post-training re-initialization and re-randomization.
Another work~\cite{DBLP:conf/nips/FrommPP18} showed that a heterogeneously quantized versions of modern networks with the right mix of different bitwidths can match the accuracy of homogeneous versions with lower effective bitwidth on average.
All aforementioned points poses a requirement for methods to efficiently discover heterogenous bitwidths assignment for neural networks.
However, exploiting this possibility is manually laborious~\cite{Micikevicius2017MixedPT, Mishra2017ApprenticeUK,DBLP:journals/corr/abs-1812-00090} as to preserve accuracy, the bitwidth varies across individual layers and different DNNs~\cite{Zhou2016DoReFaNetTL, Zhu2016TrainedTQ, Li2016TernaryWN, Mishra2017WRPNWR}.
Next subsection describes how to formulate this problem as a multi-objective optimization solved through Reinforcement Learning.

\subsection{Multi-Objective Optimization} 
Figure \ref{fig:intro} shows a sketch of the multi-objective optimization problem of layer-wise quantization of a neural network showing the underlying search space and the different design components.
Given a particular network architecture, different patterns of layer-wise quantization bitwidths form a network specific design space (possible solutions).
Pareto frontier ({\larger[1.5]\ding{182}}) defines the optimum patterns of layer-wise quantization bitwidths (Pareto optimal solutions). 
%
However, not all Pareto optimal solutions are equally of interest as the accuracy objective has a priority over the quantization objective.
In the context of neural network quantization, preferred/desired region on the Pareto frontier for a given neural network is motivated by how recoverable the accuracy is for a given state of quantization (i.e., a particular pattern of layer-wise bitwidths) of the network.
Practically, the amount of recoverable accuracy loss (upon quantization) is determined by many factors: (1) the quantized training algorithm; (2) the amount of finetuning epochs; (3) the particular set of used hyperparameters; (4) the amenability of the network to recover accuracy, which is determined by the network architecture and how much overparameterization it exhibits.
As such, a constraint ({\larger[1.5]\ding{183}}) is resulted below which even solutions on Pareto frontier are not interesting as they yield either unrecoverable or unaffordable accuracy loss depending on the application in hand.
We formulate this as a Reinforcement Learning (RL) problem where, by tuning a parametric reward function ({\larger[1.5]\ding{184}}), the RL agent can be guided towards solutions around the desirable region ({\larger[1.5]\ding{185}}) that strike a particular balance between the \bench{``State of Relative Accuracy''} (y-axis) and the \bench{``State of Quantization''} (x-axis).
As such, \releq is an automated method for efficient exploration of large hyper-parameter space (heterogeneous bitwidths assignments) that is orthogonal to the underlying quantized training technique, quantization method, network architecture, and the particular hardware platform.
Changing one or more of these components could yield different search spaces, hence, different results.

\subsection{Method Overview} 
\releq trains a reinforcement learning agent that determines the level of deep quantization (below 8 bits) for each layer of the network.
\releq agent explores the search space of the quantization levels (bit width), layer by layer.
To account for the interplay between the layers with respect to quantization and accuracy, the state space designed for \releq comprises of both static information about the layers and dynamic information regarding the network state during the RL process (Section~\ref{sec:state}).
In order to consider the effects of previous layers' quantization levels, the agent steps sequentially through the layers and chooses a  bitwidth from a predefined set, e.g., $\{2, 3, 4, 5, 6, 7, 8\}$, one layer at a time (Section~\ref{sec:action}).
The agent, consequently, receives a reward signal that is proportional to its accuracy after quantization and its benefits in term of computation and memory cost.
The underlying optimization problem is multi-objective (higher accuracy, lower compute, and reduced memory); however, preserving the accuracy is the primary concern.
To this end, we shape the reward asymmetrically to incentivize accuracy over the benefits (Section~\ref{sec:reward}).
With this formulation of the RL problem, \releq employs the state-of-the-art Proximal Policy Optimization (PPO)~\cite{ppo} to train its policy and value networks.
This section details the components and the research path we have examined to design them.

\subsection{State Space Embedding to Consider Interplay between Layers}
\label{sec:state}
\vspace{-1ex}

\renewcommand{\arraystretch}{1.2}
\newcolumntype{Y}{>{\centering\arraybackslash}X}
\begin{table}[htbp]
\centering
\small
\caption{Layer and network parameters for state embedding.}
\begin{center}
\begin{tabularx}{\columnwidth}{YYY}
\toprule
& \bf Layer Specific & \bf Network Specific \\
\midrule
\multirow{4}{*}{\bf Static} & Layer index & \multirow{4}{*}{N/A} \\
\cline{2-2} 
& Layer Dimensions & \\
\cline{2-2} 
& \makecell {Weight Statistics \\ (standard deviation)} & \\
\midrule
\multirow{3}{*}{\bf Dynamic} &  \multirow{3}{*} {\makecell{Quantization Level \\ (Bitwidth)}} & State of Quantization \\
\cline{3-3} 
&& State of Accuracy \\
\bottomrule
\end{tabularx}
\label{table:state_embed}
\end{center}
\end{table}
%
%

\niparagraph{Interplay between layers.} 
The final accuracy of a DNN is the result of interplay between its layers and reducing the bitwidth of one layer can impact how much another layer can be quantized.
Moreover, the sensitivity of accuracy varies across layers.
%
%
We design the state space and the actions to consider these sensitivities and interplay by including the knowledge about the bitwidth of previous layers, the index of the layer-under-quantization, layer size, and statistics (e.g., standard deviation) about the distribution of the weights.
However, this information is incomplete without knowing the accuracy of the network given a set of quantization levels and state of quantization for the entire network.
Table~\ref{table:state_embed} shows the parameters used to embed the state space of \releq agent, which are categorized across two different axes.
(1) \bench{``Layer-Specific''} parameters which are unique to the layer vs. \bench{``Network-Specific''} parameters that characterize the entire network as the agent steps forward during training process.
(2) \bench{``Static''} parameters that do not change during the training process vs. \bench{``Dynamic''} parameters that change during training depending on the actions taken by the agent while it explores the search space. 

\vspace{0.2cm}
\niparagraph{State of quantization and relative accuracy.}
The \bench{``Network-Specific''} parameters reflect some indication of the state of quantization and relative accuracy.
\bench{State of Quantization} is a metric to evaluate the benefit of quantization for the network and it is calculated using the compute cost and memory cost of each layer.
For a neural network with $L$ layers, we define compute cost of layer $l$ as the number of \textit{M}ultiply-\textit{Acc}umulate ($MAcc$) operations ($n^{MAcc}_l$), where ($l = 0, ..., L$).
Additionally, since \releq only quantizes weights, we define memory cost of layer $l$ as the number of weights ($n^w_l$) scaled by the ratio of \textit{Memory Access Energy} ($E_{MemoryAccess}$) to \textit{MAcc Computation Energy} ($E_{MAcc}$), which is estimated to be around $120\times$ \cite{TETRIS}. 
It is intuitive to consider the that sum of memory and compute costs linearly scales with the number of bits for each layer ($n^{bits}_l$). 
The $n^{bits}_{max}$ term is the maximum bitwidth among the predefined set of bitwidths that's available for the RL agent to pick from. 
Lastly, the \bench{State of Quantization} ($State_{Quantization}$) is the sum over all layers ($L$) that accounts for the total memory and compute cost of the entire network.
\begin{small}
\vspace{-2ex}
\[ State_{Quantization} = \frac{\sum_{l=0}^{L}\lbrack(n^w_l\times \frac{E_{MemoryAccess}}{E_{MAcc}}+n^{MAcc}_l)\times n^{bits}_l\rbrack}{\sum_{l=0}^{L}\lbrack n^w_l\times \frac{E_{MemoryAccess}}{E_{MAcc}}+n^{MAcc}_l\rbrack\times n^{bits}_{max}} \]
\end{small}%
Besides the potential benefits, captured by $State_{Quantization}$, \releq considers the \bench{State of Relative Accuracy} to gauge the effects of quantization on the classification performance.
To that end, the \bench{State of Relative Accuracy} ($State_{Accuracy}$) is defined as the ratio of the current accuracy ($Acc_{Curr}$) with the current bitwidths for all layers during RL training, to accuracy of the network when it runs with full precision ($Acc_{FullP}$).
$State_{Accuracy}$ represents the degradation of accuracy as the result of quantization. 
The closers this term is to $1.0$, the lower the accuracy loss and more desirable the quantization levels.
\begin{small}
\vspace{-1ex}
\[ State_{Accuracy} = \frac {Acc_{Curr}}{Acc_{FullP}} \]
\end{small}
\vspace{-2ex}

Given these embedding of the observations from the environment, the \releq agent can take actions, described next.

\subsection{Flexible Actions Space}
\label{sec:action}
\vspace{-1ex}

\begin{figure}
    \centering
    \includegraphics[width=0.425\textwidth]{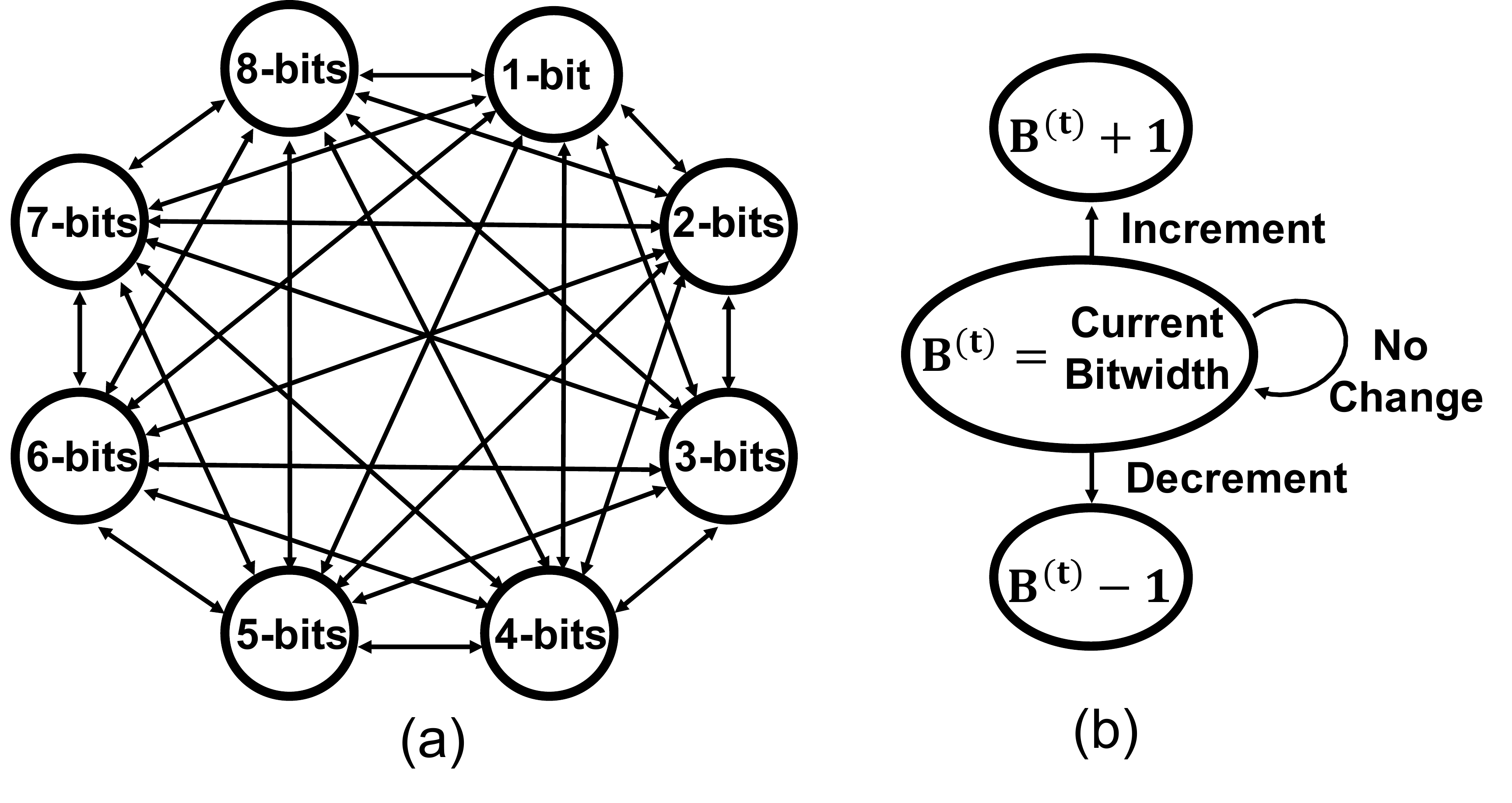}
    \caption{(a) Flexible action space (used in \releq). (b)  Alternative action space with restricted movement.}		  
    \label{fig:action_space}
    \vspace{-0.5cm}
\end{figure}

%
Intuitively, as calculations propagate through the layers, the effects of quantization will accumulate. 
As such, the \releq agents steps through each layer sequentially and chooses from the bitwidth of a layer from a discrete set of quantization levels which are provided as possible choices.
Figure~\ref{fig:action_space}(a) shows the representation of action space in which the set of bitwidths is $\{1, 2, 3, 4, 5, 6, 7, 8\}$. 
As depicted, the agent can flexibly choose to change the quantization level of a given layer from any bitwidth to any other bitwidth.
The set of possibilities can be changed as desired. 
Nonetheless, the action space depicted in Figure~\ref{fig:action_space}(a) is the possibilities considered for deep quantization in this work.
As illustrated in Figure~\ref{fig:action_space}(b), an alternative that we experimented with was to only allow the \releq agent to increment/decrement/keep the current bitwidth of the layer ($B^{(t)}$).
The experimentation showed that the convergence is much longer than the aforementioned flexible action space, which is used. 

\begin{figure}
    \centering
    \includegraphics[width=0.52\textwidth]{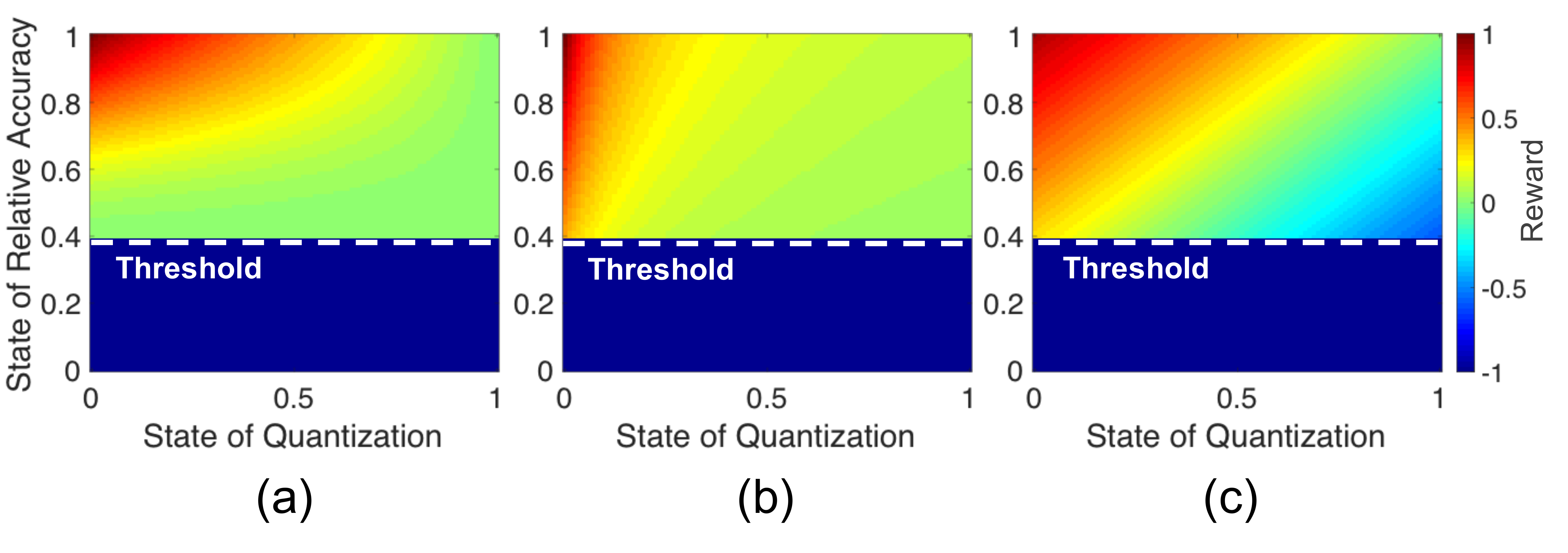}
    \caption{Reward shaping with three different formulations.	The color palette shows the intensity of the reward.}
    \label{fig:reward_shapes}
    \vspace{-0.5cm}
\end{figure}

\subsection{Asymmetric Reward Formulation for Accuracy}
\label{sec:reward}
\vspace{-1ex}

While the state space embedding focused on interplay between the layers and the action space provided flexibility, reward formulation for \releq aims to preserve accuracy and minimize bitwidth of the layers simultaneously.
This requirement creates an asymmetry between the accuracy and bitwidth reduction, which is a core objective of \releq.
The following \bench{Reward Shaping} formulation provides the asymmetry and puts more emphasis on  maintaining the accuracy as illustrated with different color intensities in Figure~\ref{fig:reward_shapes}(a).
%
\input{listings/algorithm.tex}
%
This reward uses the same terms of \bench{State of Quantization} ($State_{Quantization}$) and \bench{State of Relative Accuracy} ($State_{Accuracy}$) from Section~\ref{sec:state}.
One of the reasons that we chose this formulation is that it produces a smooth reward gradient as the agent approaches the optimum quantization combination.
In addition, the varying 2-dimensional gradient speeds up the agent's convergence time.
In the reward formulation, $a=0.2$ and $b=0.4$ can also be tuned and $th=0.4$ is threshold for relative accuracy below which the accuracy loss may not be recoverable and those quantization levels are completely unacceptable.
The use of threshold also accelerates learning as it prevents unnecessary or undesirable exploration in the search space by penalizing the agent when it explores undesired low-accuracy states.
While Figure~\ref{fig:reward_shapes}(a) shows the aforementioned formulation, Figures~\ref{fig:reward_shapes}(b) and (c) depict two other alternatives. 
Figure~\ref{fig:reward_shapes}(b) is based on $State_{Accuracy}/State_{Quantization}$ while Figure~\ref{fig:reward_shapes}(c) is based on  $State_{Accuracy} - State_{Quantization}$.
Section~\ref{sec:rewardsens} provides detailed experimental results with these three reward formulations.
In summary, the formulation for Figure~\ref{fig:reward_shapes}(a) offers faster convergence. 
%
%
\begin{figure*}
    \centering
    \includegraphics[width=0.9\textwidth]{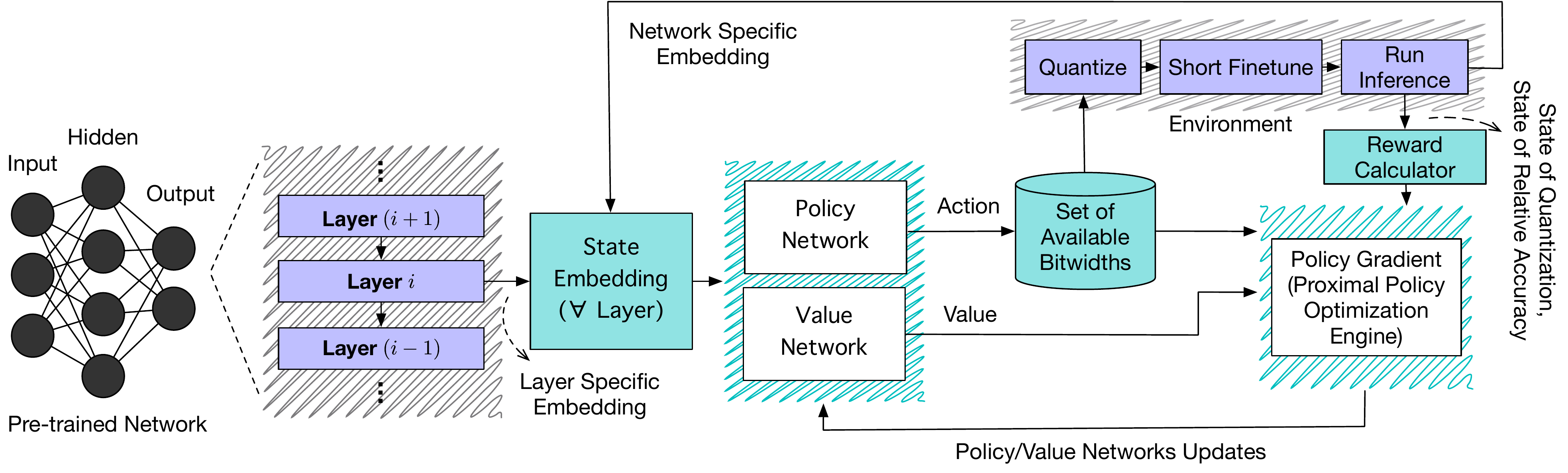}
    \caption{Overview of \releq, which starts from a pre-trained network and delivers its corresponding deeply quantized network.}
    \label{fig:overview}
    \vspace{-0.5cm}
\end{figure*}
\begin{figure}
  	  \centering\includegraphics[width=0.45\textwidth]{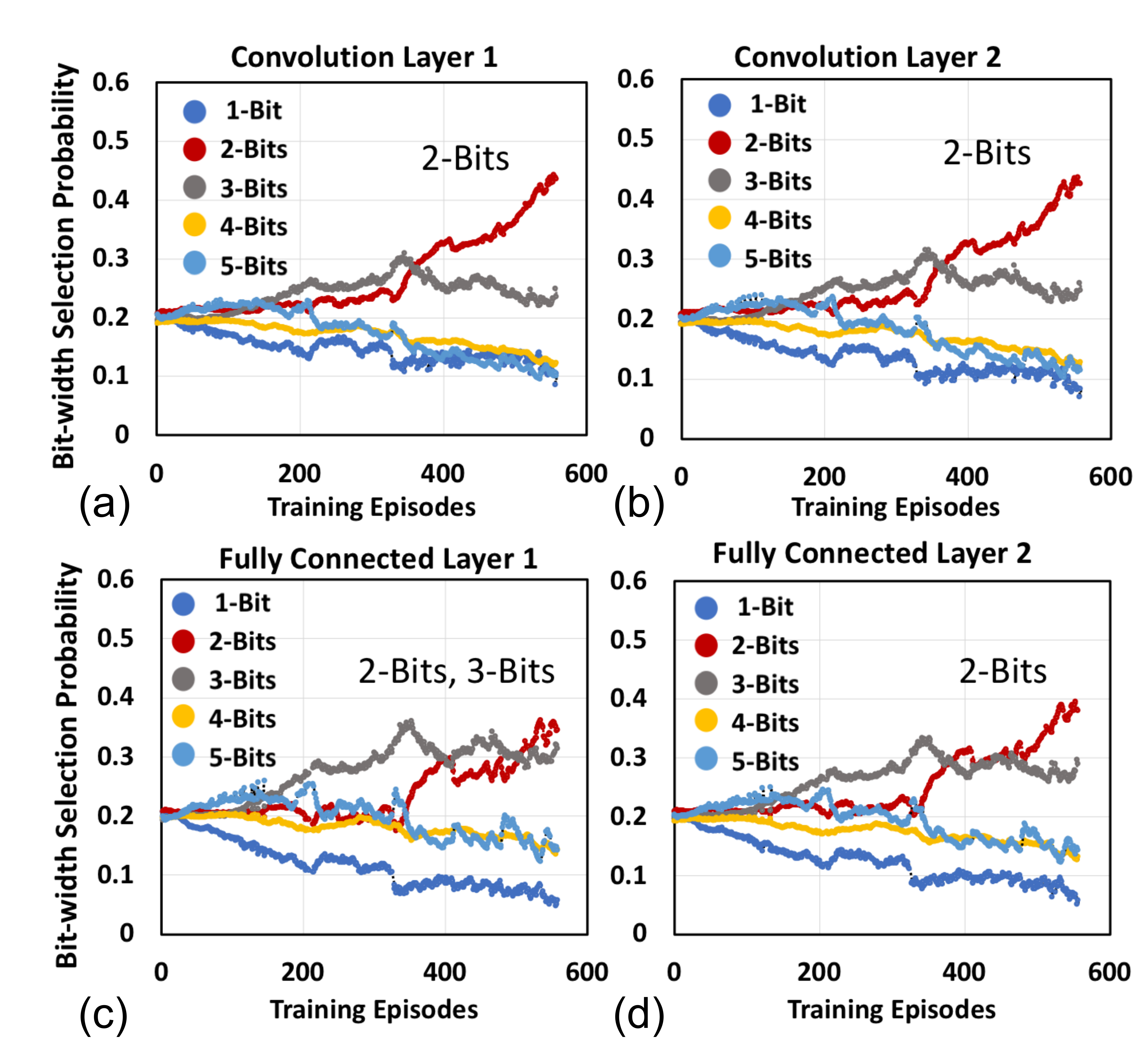}
	  \caption{Action (Bitwidths selection) probability evolution over training episodes for LeNet.}
	  \label{fig:releq_lenet_policy}
	  \vspace{-0.5cm}
 \end{figure}
\subsection{Policy and Value Networks}
\vspace{-1ex}
While state, action and reward are three essential components of any RL problem, Policy and Value complete the puzzle and encode learning in terms of a RL context. 
While there are both policy-based and value-based learning techniques, \releq uses a state-of-the-art policy gradient based approach, Proximal Policy Optimization (PPO)~\cite{ppo}.
PPO is a actor critic style algorithm so \releq agent consists of both Policy and Value networks.
%

\vspace{0.2cm}
\niparagraph{Network architecture of Policy and Value networks.}
Both Policy and Value are functions of state, so the the state space defined in Section~\ref{sec:state} is encoded as a vector and fed as input to a Long short-term memory(LSTM) layer and this acts as the first hidden layer for both policy and value networks. 
Apart from the LSTM, policy network has two fully connected hidden layers of 128 neurons each and the number of neurons in the final output layer is equal to the number of available bitwidths the agent can choose from. Whereas the Value network has two fully connected hidden layers of 128 and 64 neurons each. 
Based on our evaluations, LSTM enables the \releq agent to converge almost $\times{1.33}$ faster in comparison to a network with only fully connected layers.

While this section focused on describing the components of \releq in isolation, the next section puts them together and shows how \releq automatically quantizes a pre-trained DNN.

%% file: listings/algorithm.tex
\begin{algorithm}
\begin{algorithmic}
  \algsetup{linenosize=\tiny}
\STATE \textbf{\underline{Reward Shaping:}}
\STATE $ reward = 1.0 - (State_{Quantization})^a $ 
\IF{$(State_{Acc} < th) $}
	\STATE {$reward = -1.0 $}
\ELSE
	\STATE $Acc_{discount} = max(State_{Acc}, th)^{(b/max(State_{Acc}, th))}$ 
	\STATE $reward = reward \times Acc_{discount}$ 
\ENDIF 
\end{algorithmic}
\end{algorithm}

%% file: body/learning_proc.tex
 
\section{Putting it All Together: \releq in Action}
\label{sec:releq_working}
\vspace{-1ex}

As discussed in Section~\ref{sec:rl_quantization}, state, action and reward enable the \releq agent to maneuver the search space with an objective of quantizing the neural network with minimal loss in accuracy.
\releq starts with a pretrained model of full precision weights and proposes quantization levels of weights for all layers in a DNN.
Figure~\ref{fig:overview} depicts the entire workflow for \releq and this section gives an overview of how everything fits together.

\vspace{0.2cm}
\niparagraph{Interacting with the environment.}
\releq agent steps through all layers one by one, determining the quantization level for the layer at each step.
For every step, the state embedding for the current layer comprising of different elements described in Section~\ref{sec:state} is fed as an input to the Policy and Value Networks of the \releq agent and the output is the probability distribution over the different possible bitwidths and value of the state respectively.
\releq agent then takes a stochastic action based on this probablity distribution and chooses a quantization level for the current layer.
Weights for this particular layer are quantized to the predicted bitwidth and  with accuracy preservation being a primary component of \releq's reward function, retraining of a quantized neural network is required in order to properly evaluate the effectiveness of deep quantization.
Such retraining is a time-intensive process and it undermines the search process efficiency.
To get around this issue, we reward the agent with an estimated validation accuracy after retraining for a shortened amount of epochs.
For deeper networks, however, due to the longer retraining time, we perform this phase after all the bitwidths are selected for the layers and then we do a short retraining.
Dynamic network specific parameters listed in Table~\ref{table:state_embed} are updated based on the validation accuracy and current quantization levels of the entire network before stepping on to the next layer.
In this context, we define an epsiode as a single pass through the entire neural network and the end of every episode, we use Proximal Policy Optimization~\cite{ppo} to update the Policy and Value networks of the \releq agent.
After the learning process is complete and the agent has converged to a quantization level for each layer of the network, for example 2 bits for second layer, 3 bits for fourth layer and so on, we perform a long retraining step using the quantized bitwidths predicted by the agent and then obtain the final accuracy for the quantized version of the network.
%
%
%
%
%
\vspace{0.2cm}
\niparagraph{Learning the Policy.}
Policy in terms of neural network quantization is to learn to choose the optimal bitwidth for each layer in the network.
Since \releq uses a Policy Gradient based approach, the objective is to optimize the policy directly, it's possible to visualize how policy for each layer evolves with respect to time i.e the number of episodes.
Figure \ref{fig:releq_lenet_policy} shows the evolution of \releq agent's bitwidth selection probabilities for all layers over time (number of episodes), which reveals how the agent's policy changes with respect to selecting a bitwidth per layer. As indicated on the graph, the end results suggest the following quantization patterns, ${2, 2, 2, 2}$ or ${2, 2, 3, 2}$ bits. 
For the first two convolution layers (Convolution Layer 1, Convolution Layer 2), the agent ends up assigning the highest probability for two bits and its confidence increases with increasing number of training episodes.
For the third layer (Fully Connected Layer 1), the probabilities of two bits and three bits are very close. Lastly, for the fourth layer (Fully Connected Layer 2), the agent again tends to select two bits, however, with relatively smaller confidence compared to layers one and two.
With these observations, we can infer that bitwidth probability profiles are not uniform across all layers and that the agent distinguishes between the layers, understands the sensitivity of the objective function to the different layers and accordingly chooses the bitwidths.
%
%
Looking at the agent's selection for the third layer (Fully Connected Layer 1) and recalling the initial problem formulation of quantizing all layers while preserving the initial full precision accuracy, it is logical that the probabilities for two and three bits are very close. Going further down to two bits was beneficial in terms of quantization while staying at three bits was better for maintaining good accuracy which implies that third layer precision affects accuracy the most for this specific network architecture. \textit{This points out the importance of tailoring the reward function and the role it plays in controlling optimization tradeoffs}.

%% file: body/experimental_setup_new.tex
\section{Experimental Setup}
\label{sec:exp}
%

\newcolumntype{Y}{>{\centering\arraybackslash}X}
\begin{table*}
\small
    \centering
	\caption{Benchmark DNNs and their deep quantization with \releq.}
\begin{tabularx}{\textwidth}{lcccc@{}}
\toprule
\bf Network     & \bf Dataset & \bf Quantization Bitwidths & \bf Average Bitwidth & \bf \makecell{Acc Loss (\%)} \\ 
\midrule
AlexNet& ImageNet & \bf \{8, 4, 4, 4, 4, 4, 4, 8\} & 5 & \bf 0.08 \\
\addlinespace
SimpleNet& CIFAR10 & \bf \{5, 5, 5, 5, 5\} & 5 &\bf 0.30 \\
\addlinespace
LeNet& MNIST & \bf \{2, 2, 3, 2\} & 2.25 & \bf0.00 \\
\addlinespace
MobileNet& ImageNet & \makecell{\bf \{8, 5, 6, 6, 4, 4, 7, 8, 4, 6, 8, 5, 5, 8, 6, 7, 7, 7, 6, 8, 6, 8, 8, 6, 7, 5, 5, 7, 8, 8\}} & 6.43 & \bf0.26 \\
\addlinespace
ResNet-20& CIFAR10 & \bf {\{8, 2, 2, 3, 2, 2, 2, 3, 2, 3, 3, 3, 2, 2, 2, 3, 2, 2, 2, 2, 2, 8\}} & 2.81 & \bf0.12 \\
\addlinespace
SVHN-10& SVHN & \bf \{8, 4, 4, 4, 4, 4, 4, 4, 4, 8\} & 4.80 & \bf0.00 \\
\addlinespace
VGG-11& CIFAR10 & \bf \{8, 5, 8, 5, 6, 6, 6, 6, 8\} & 6.44 & \bf0.17 \\
\addlinespace
VGG-16& CIFAR10 & \bf \{8, 8, 8, 6, 8, 6, 8, 6, 8, 6, 8, 6, 8, 6, 8, 8\} & 7.25 & \bf0.10 \\
\bottomrule
\end{tabularx}
\label{table:main_results}
\end{table*}

%
\subsection{Benchmarks}
To assess the effectiveness of \releq across a variety of DNNs, we use the following seven diverse networks that have been used in different real-world vision tasks: AlexNet, CIFAR-10 (Simplenet), LeNet, MobileNet (Version 1), ResNet-20, SVHN and VGG-11.
Of these seven networks, AlexNet and MobileNet were evaluated on the ImageNet (ILSVRC'12) dataset, ResNet-20, VGG-11 and SimpleNet (5 layers) on CIFAR-10, SVHN (10 layers) on SVHN  and LeNet on the MNIST dataset.  

\subsection{Quantization Technique} 
As described in earlier sections, \releq is an off-the-shelf automated framework that works on top of any existing quantization technique to yield efficient heterogeneous bitwidths assignments. Here, we use the technique proposed in WRPN \cite{Mishra2017WRPNWR} where weights are first scaled and clipped to the $(-1.0, 1.0)$ range and quantized as per the following equation. 
The parameter $k$ is the bitwidth used for quantization out of which $k-1$ bits are used for quantization and one bit is used for sign.
%
%

\begin{equation}
w_{q} = \frac{round((2^{k-1} - 1)w_{f})}{2^{k-1} - 1}
\end{equation}
%
Additionally, different quantization styles (e.g., mid-tread vs. mid-rise) yield different quantization levels. In mid-tread, zero is considered as a quantization level, while in mid-rise, quantization levels are shifted by half a step such that zero is not included as a quantization level. Here, we use mid-tread style following WRPN.
\subsection{Granularity of Quantization}
Quantization can come at different granularities: per-network, per-layer, per-channel/group, or per-parameter~\cite{DBLP:journals/corr/abs-1806-08342}. However, as the granularity becomes finer, the search space goes exponentially larger. Here, we consider per-layer granularity which strikes a balance between adapting for specific network requirements and practical implementations as supported in a wide range of hardware platforms such as CPUs, FPGAs, and dedicated accelerators. Nonetheless, similar principles of automated optimization can be extended for other granularities as needed. 
\subsection{Deep Quantization with Conventional Hardware} \releq's solution can be deployed on conventional hardware, such as general purpose CPUs to provide benefits and improvements. 
To manifest this, we have evaluated \releq using TVM~\cite{TVM} on an Intel Core i7-4790 CPU.
We use TVM since its compiler supports deeply quantized operations with bit-serial vector operations on conventional hardware. 
We compare our solution in terms of the inference execution time (since the TVM framework does not offer energy measurements) against 8-bit quantized network. 
The results can be seen in Figure~\ref{fig:tvm} and will be further elaborated in the next section.

\subsection{Deep Quantization with Custom Hardware Accelerators} To further demonstrate the energy and performance benefits of the solution found by \releq, we evaluate it on Stripes~\cite{stripes}, a custom accelerator designed for DNNs, which exploits bit-serial computation to support flexible bitwidths for DNN operations.
Stripes does not support or benefit from deep quantization of activations and it only leverages the quantization of weights. 
We compare our solution in terms of energy consumed and inference execution time against the 8-bit quantized network.

\subsection{Comparison with Prior Work} 
We also compare against prior work~\cite{Ye2018AUF}, which proposes an iterative optimization procedure (dubbed ADMM) through which they find quantization bitwidths only for AlexNet and LeNet. 
Using Stripes~\cite{stripes} and TVM~\cite{TVM}, we show that \releq's solution provides higher performance and energy benefits compared to ADMM~\cite{Ye2018AUF}. 
%

\begin{table}
    \centering
	\caption{Hyperparameters of PPO used in ReLeQ.}
\begin{tabularx}{\columnwidth}{lY@{}}
\toprule
\bf Hyperparameter     & \bf Value \\ 
\midrule
Adam Step Size & $1\times10^{-4}$\\
Generalized Advantage Estimation Parameter & $0.99$\\
Number of Epochs & $3$\\
Clipping Parameter & $0.1$\\
\bottomrule
\end{tabularx}
\label{table:ppo_params}
\end{table}

\subsection{Implementation and Hyper-parameters of the Proximal Policy Optimization (PPO)}
As discussed, \releq uses PPO~\cite{ppo} as its RL engine, which we implemented in python where its policy and value networks use TensorFlow's Adam Optimizer with an initial learning rate of $10^{-4}$. 
The setting of the other hyper-parameters of PPO is listed in Table~\ref{table:ppo_params}.

%% file: body/evaluation_new.tex
\section{Experimental Results}
%
%

\begin{figure}
\centering 	
 	\includegraphics[width=0.45\textwidth]{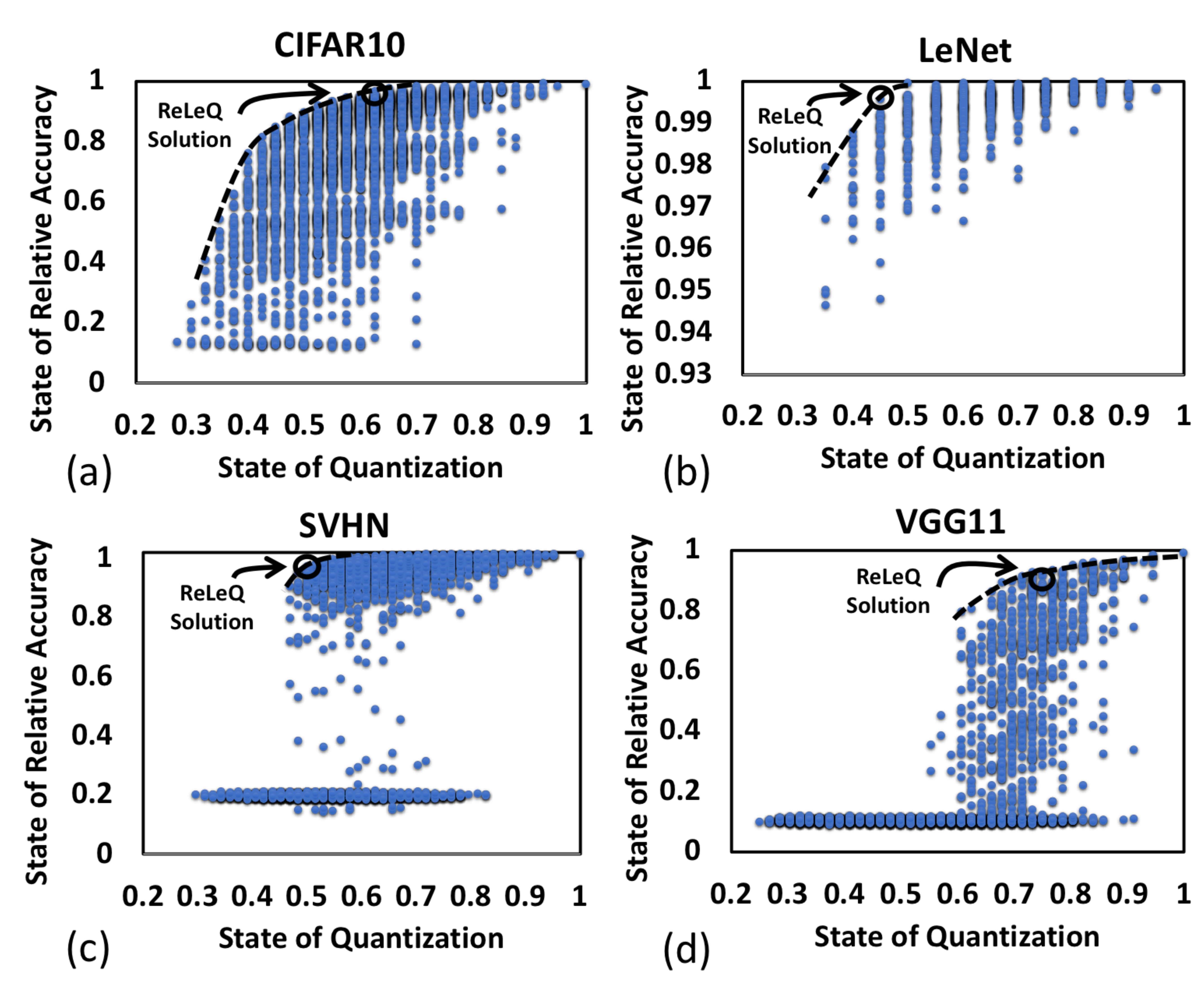}
	\caption{Quantization space and its Pareto frontier for  
	(a) CIFAR-10, (b) LeNet, (c) SVHN, and (d) VGG-11.}
	\label{fig:enum}
	\vspace{-0.5cm}
 \end{figure}

\subsection{Quantization Levels with \releq}\label{sec:quant_policy}
Table~\ref{table:main_results} provides a summary of the evaluated networks, datasets and shows the results with respect to layer-wise quantization levels (bitwidths) achieved by \releq. 
%
%
Regarding the layer-wise quantization bitwidths, at the onset of the agent's exploration, all layers are initialized to 8-bits.
As the agent learns the optimal policy, each layer converges with a high probability to a particular quantization bitwidth. 
As shown in the \bench{``Quantization Bitwidths''} column of  Table~\ref{table:main_results}, \releq quantization policies show a spectrum of varying bitwidth assignments to the layers.
The bitwidth for MobileNet varies from 4 bits to 8 bits with an irregular pattern, which averages to 6.43.
ResNet-20 achieves mostly 2 and 3 bits, again with an irregular interleaving that averages to 2.81.
In many cases, there is significant heterogeneity and irregularity in the bitwidths and a uniform assignment of the bits is not always the desired choice to preserve accuracy.
These results demonstrate that ReLeQ automatically distinguishes different layers and their varying importance with respect to accuracy while choosing their respective bitwidths.
As shown in the \bench{``Accuracy Loss''} column of  Table~\ref{table:main_results}, the deeply quantized networks with \releq have less than 0.30\% loss in classification accuracy. 
%
%
To assess the quality of these bitwidths assignments, we conduct a Pareto analysis on the DNNs for which we could populate the search space.

\begin{figure*}
\centering
    \includegraphics[width=0.8\textwidth]{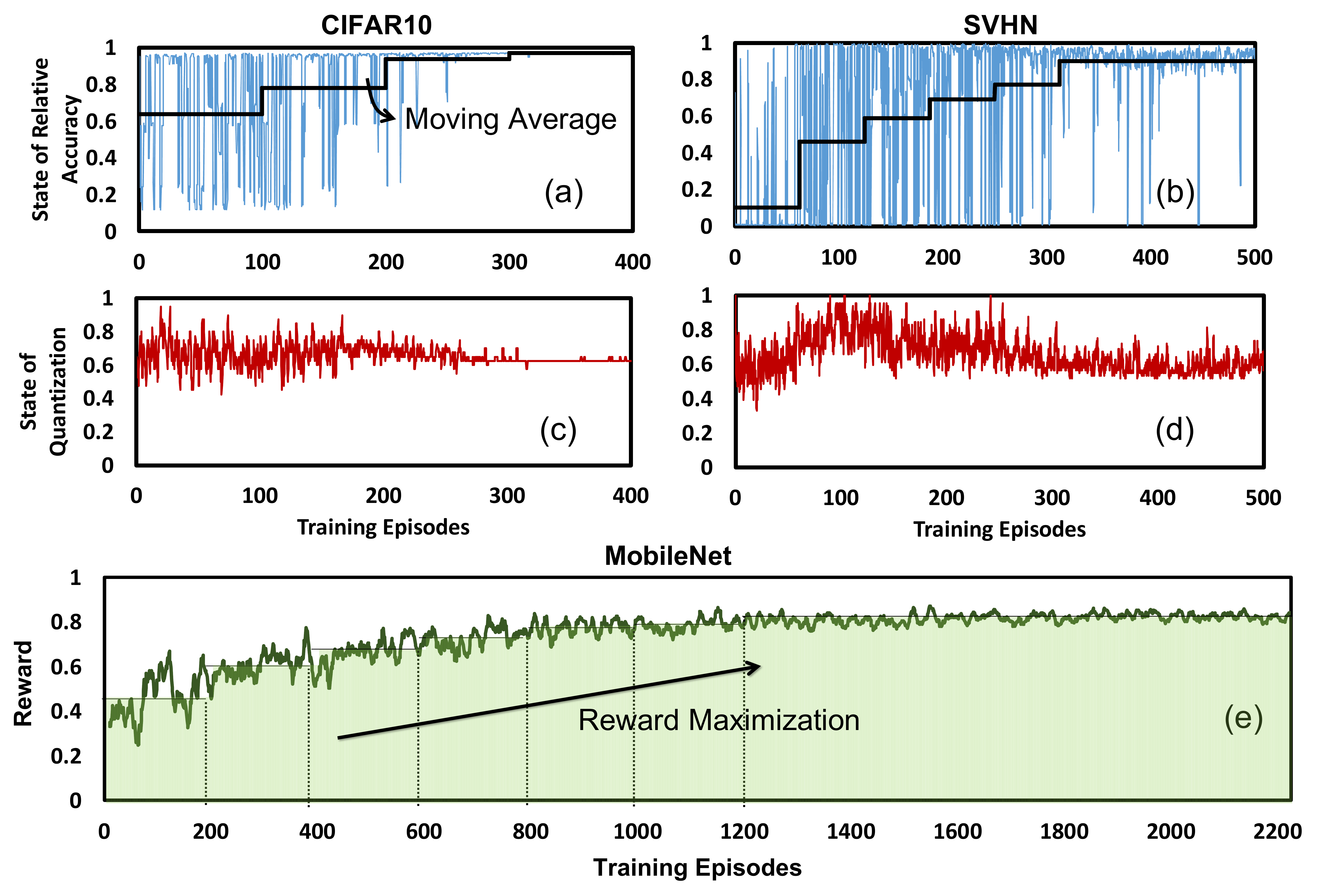}
    \caption{The evolution of reward and its basic elements: State of Relative Accuracy for (a) CIFAR-10, (b) SVHN.  State of Quantization for (c) CIFAR-10, (d) SVHN, 		as the agent learns through the episodes. 
    		The last plot (e) shows an alternative view by depicting the evolution of reward for MobileNet.
		The trends are similar for the other networks.}
    \label{fig:convergence}
\end{figure*}

\subsection{Validation: Pareto Analysis}\label{sec:Validation}
Figure \ref{fig:enum} depicts the solutions space for four benchmarks (CIFAR10, LeNet, SVHN, and VGG11).
Each point on these charts is a unique combination of bitwidths that are assigned to the layers of the network.
The boundary of the solutions denotes the Pareto frontier and is highlighted by a dashed line.
The solution found by \releq is marked out using an arrow and lays on the desired section of the Pareto frontier where the accuracy loss can be recovered through fine-tuning, which demonstrates the quality of the obtained solutions.
It is worth noting that as a result of the moderate size of the four networks presented in this subsection, it was possible to enumerate the design space, obtain Pareto frontier and assess ReLeQ quantization policy for each of the four networks. 
However, it is infeasible to do so for state-of-the-art deep networks (e.g., MobileNet and AlexNet) which further stresses the importance of automation and efficacy of \releq.

\subsection{Learning and Convergence Analysis}\label{sec:Convergence}
We further study the desired behavior of \releq in the context of convergence.
An appropriate evidence for the correctness of a formulated reinforcement learning problem is the ability of the agent to consistently yield improved solutions.
The expectation is that the agent learns the correct underlying policy over the episodes and gradually transitions from the exploration to the exploitation phase. 
Figures~\ref{fig:convergence}(a) and (b) first show the \bench{State of Relative Accuracy} for CIFAR10 and SVHN, respectively. 
We overlay the moving average of \bench{State of Relative Accuracy} as episodes evolve, which is denoted by a black line in Figures~\ref{fig:convergence}(a) and (b).
Similarly, Figures~\ref{fig:convergence}(c) and (d) depict the evolution of \bench{State of Quantization}. 
As another indicative parameter of learning, Figure~\ref{fig:convergence}(e) plots the evolution of the reward, which combines the two \bench{States of Accuracy} and \bench{Quantization} (Section~\ref{sec:reward}).
As all the graphs show, the agent consistently yields solutions that increasingly preserve the accuracy (maximize rewards), while seeking to minimize the number of bits assigned to each layer (minimizing the state of quantization) and eventually converges to a rather stable solution. 
The trends are similar for the other networks.

%

 \begin{figure}
	\centering\includegraphics[width=0.45\textwidth]{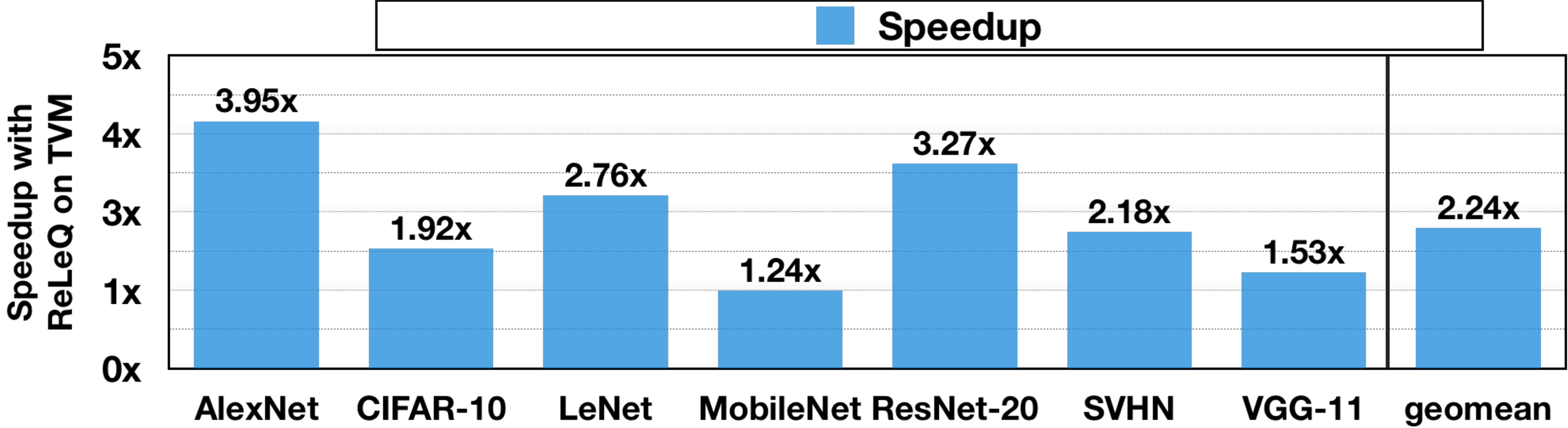}
	\caption{Speedup with \releq for conventional hardware using TVM over the baseline run using 8 bits.}
	\label{fig:tvm}
\end{figure}
	
 \begin{figure}
	\centering\includegraphics[width=0.45\textwidth]{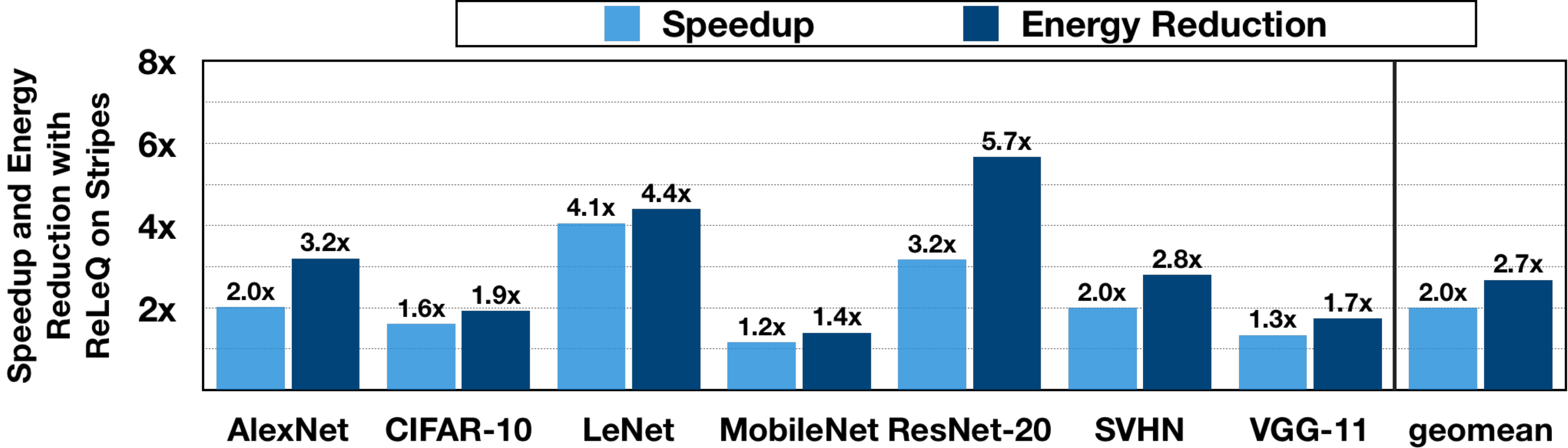}
	\caption{Energy reduction and speedup with \releq for Stripes over the baseline execution when the accelerator is running 8-bit DNNs.}
	\label{fig:stripes}
\end{figure}

\renewcommand{\arraystretch}{1.4}
\begin{table*}[htbp]
\centering
\small
\caption{Speedup and energy reduction with \releq over ADMM~\cite{Ye2018AUF}.}
\begin{center}
\begin{tabular}{lcccccc}
\toprule
{\textbf{Network}} & \textbf{{Dataset}}& \textbf{{Technique}}& \textbf{{Bitwidth}}& \textbf{{\makecell{\releq speedup \\ on TVM}}}& \textbf{{\makecell{\releq speedup \\ on Stripes}}} & \textbf{{\makecell{Energy Improvement of \\ \releq on Stripes}}} \\
\hline
\multirow{2}{*}{AlexNet} & \multirow{2}{*}{ImageNet} & \textbf{\releq} & \textbf{\{8,4,4,4,4,4,4,8\}} & \multirow{2}{*}{1.20$X$} & \multirow{2}{*}{1.22$X$} & \multirow{2}{*}{1.25$X$} \\
&& ADMM & \{8,5,5,5,5,3,3,8\} & & & \\
\hline
\multirow{2}{*}{LeNet} & \multirow{2}{*}{MNIST} & \textbf{\releq} & \textbf{\{2,2,3,2\}} & \multirow{2}{*}{1.42$X$} & \multirow{2}{*}{1.86$X$} & \multirow{2}{*}{1.87$X$} \\
&& ADMM & \{5,3,2,3\} & & & \\
\bottomrule
\end{tabular}
\label{table:res_comp}
\end{center}
\end{table*}

\begin{figure*}
  	  \centering\includegraphics[width=0.8\textwidth]{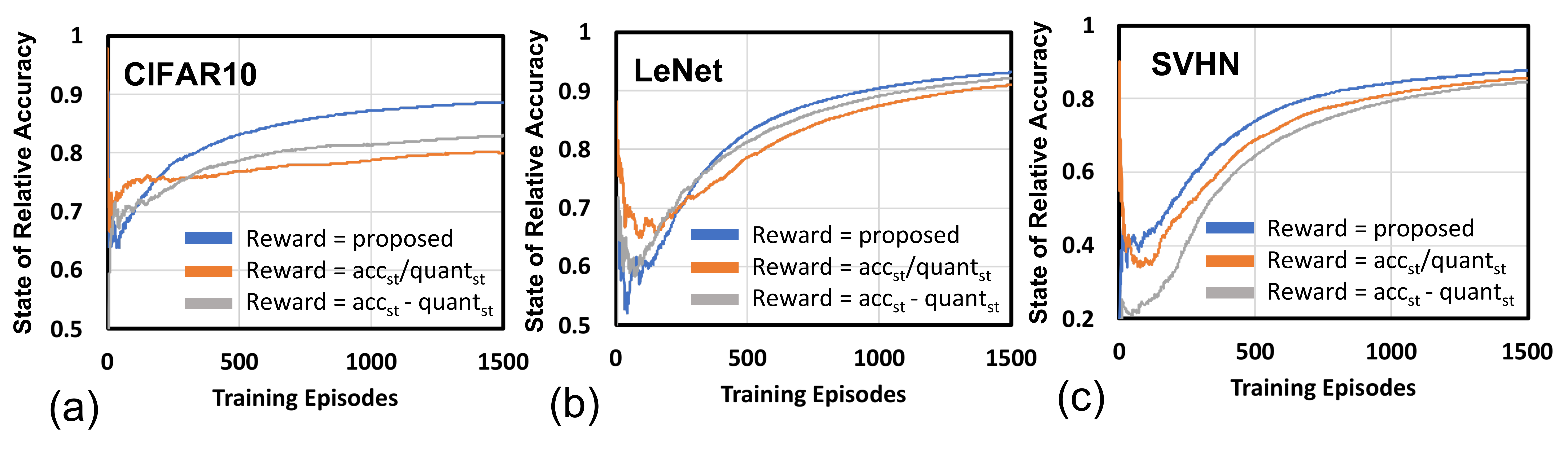}
	  \caption{Three different reward functions and their impact on the state of relative accuracy over the training episodes for three different networks. (a) CIFAR-10, (b) LeNet, and (c) SVHN.}
	  \label{fig:diff_rewards}
 \end{figure*}

\subsection{Execution Time and Energy Benefits with \releq} 
Figure~\ref{fig:tvm} shows the speedup for each benchmark network on conventional hardware using TVM compiler. 
The baseline is the 8-bit runtime for inference. 
\releq's solution offers, on average, $2.2\times$ speedup over the baseline as the result of merely quantizing the weights that reduces the amount of computation and data transfer during inference. 
Figure~\ref{fig:stripes} shows the speedup and energy reduction benefits of \releq's solution on Stripes custom accelerator. 
The baseline here is the time and energy consumption of 8-bit inference execution on the same accelerator.

\releq's solutions yield, on average, $2.0\times$ speedup and an additional $2.7\times$ energy reduction. 
MobileNet achieves $1.2\times$ speedup which is coupled with a similar degree of energy reduction.
On the other end of the spectrum, ResNet-20 and LeNet achieve $3.0\times$ and $4.0\times$ benefits, respectively.
As shown in Table~\ref{table:main_results}, MobileNet needs to be quantized to higher bitwidths to maintain accuracy, compared with other networks and that is why the benefits are smaller. 

\subsection{Speedup and Energy Reduction over ADMM} 
As mentioned in Section~\ref{sec:exp}, we compare \releq's solution in terms of speedup and energy reduction against ADMM~\cite{Ye2018AUF}, another procedure for finding quantization bitwidths. 
As shown in Table~\ref{table:res_comp}, \releq's solution provides $1.25\times$ energy reduction and $1.22\times$ average speedup over ADMM with Stripes for AlexNet and the benefits are higher for LeNet.
The benefits are similar for the conventional hardware using TVM as shown in Table~\ref{table:res_comp}.
ADMM does not report other networks.

\subsection{Sensitivity Analysis: Influence of Reward Function}\label{sec:rewardsens}
The design of reward function is a crucial component of reinforcement learning as indicated in Section~\ref{sec:reward}.
There are many possible reward functions one could define for a particular application setting. However, different designs could lead to either different policies or different convergence behaviors.
In this paper, we incorporate reward engineering by proposing a special parametric reward formulation.
To evaluate the effectiveness of the proposed reward formulation, we have compared three different reward formulations in Figure \ref{fig:diff_rewards}: (a) proposed in Section~\ref{sec:reward}, (b) $ R=State_{Accuracy}/State_{Quantization}$, (c) $ R=State_{Accuracy}-State_{Quantization}$.
As the blue line in all the charts shows, the proposed reward formulation consistently achieves higher \bench{State of Relative Accuracy} during the training episodes.
That is, our proposed reward formulation enables \releq finds better solutions in shorter time.

\subsection{Tuning: PPO Objective Clipping Parameter}\label{sec:clipping_param_tuning}
One of the unique features about PPO algorithm is its novel objective function with clipped probability ratios, which forms a lower-bound estimate of the change in policy. 
Such modification controls the variance of the new policy from the old one, hence, improves the stability of the learning process.
PPO uses a Clipped Surrogate Objective function, which uses the minimum of two probability ratios, one non-clipped and one clipped in a range between $\lbrack1-\epsilon$, $1+\epsilon\rbrack$, where $\epsilon$ is a hyper-parameter that helps to define this clipping range. 
Table \ref{table:clipping_table} provides a summary of tuning epsilon (commonly in the range of 0.1, 0.2, 0.3). Based on our performed experiments, $\epsilon=0.1$ often reports the highest average reward across different benchmarks.

\renewcommand{\arraystretch}{1.4}
\begin{table}[htbp]
\centering
\small
\caption{Sensitivity of reward to different clipping parameters.}
\begin{center}
\begin{tabularx}{\columnwidth}{YYYY}
\toprule
\multirow{2}{*}{\makecell{\bf PPO Clipping \\ \bf Parameter}} &  \multicolumn{3}{{c}}{\bf Average Normalized Reward (Performance) } \\
\cline{2-4}
 & \bf LeNet on MNIST & \bf SimpleNet on CIFAR-10 & \bf 8-Layers on SVHN \\
\midrule
 $\epsilon = 0.1$ & \bf 0.209 & 0.407 & \bf 0.499 \\
 $\epsilon = 0.2$ & 0.165 & \bf 0.411 & 0.477 \\
 $\epsilon = 0.3$ & 0.160 & 0.399 & 0.455 \\
\bottomrule
\end{tabularx}
\label{table:clipping_table}
\end{center}
\end{table}

%% file: body/related_new.tex
\section{Related Work}
%
ReLeQ is the initial step in utilizing reinforcement learning to automatically find the level of quantization for the layers of DNNs such that their classification accuracy is preserved.
As such, it relates to the techniques that given a level of quantization, train a neural network or develop binarized DNNs.
Furthermore, the line of research that utilizes RL for hyperparameter discovery and tuning inspires \releq.
Nonetheless, \releq, uniquely and exclusively, offers an RL-based approach to determine the levels of quantization.

\vspace{0.1cm}
\niparagraph{Automated machine learning (AutoML) methods.}
Because of the increased deployment of deep learning models into various domains and applications, AutoML has recently gained a substantial interest from both academia~\cite{DBLP:books/sp/19/FeurerKES0H19, DBLP:books/sp/19/MendozaKFSUBDLH19} and industry as internal tools~\cite{DBLP:conf/kdd/GolovinSMKKS17}, or open services~\cite{google_cloud_autoML, amazon_autoML}. 
Hyperparameter optimization (HPO) is a major subfield of AutoML. 
One notable AutoML system is Auto-sklearn~\cite{DBLP:books/sp/19/FeurerKES0H19} that uses Bayesian optimization method to find the best instantiation of classifiers in scikit-learn~\cite{scikit-learn}.


%

%

%
Another major application of AutoML is neural architecture search (NAS) which also a has significant overlap with hyperparameter optimization~\cite{DBLP:journals/jmlr/ElskenMH19}.
Recently, Google introduced Cloud AutoML service~\cite{google_cloud_autoML} that is suite of machine learning products that enables developers with limited machine learning expertise to train high-quality models specific to their business needs. It relies on Google’s state-of-the-art transfer learning and neural architecture search technology.
Even though most of these frameworks include a wide range of supervised learning methods, a little include modern neural networks and their optimization.

\vspace{0.1cm}
\niparagraph{Reinforcement learning for automatic tuning.}
%
%
RL based methods have attracted much attention within NAS after obtaining the competitive performance on the CIFAR-10 dataset employing RL as the search strategy despite the massive amount of the used computational resources~\cite{Zoph2016NeuralAS}. 
Different RL approaches differ in how they represent the agent’s policy. Zoph and Le~\cite{Zoph2016NeuralAS} use a recurrent neural network (RNN) trained by policy gradient, in particular, REINFORCE, to sequentially sample a string that in turn encodes a neural architecture. 
Baker et. al.~\cite{Baker2016DesigningNN} use Q-learning to train a policy which sequentially chooses a layer’s type and corresponding hyperparameters.
%

%
Recently, a wide variety of methods have been proposed in quick succession to reduce the computational costs and achieve further performance improvements~\cite{ DBLP:conf/icml/YingKCR0H19, DBLP:conf/icml/PhamGZLD18}.

Aside from NAS applications, i.e. engineering neural architecture from scratch, \cite{He2018AMCAF} employ RL to prune existing architectures where a policy gradient method is used to automatically find the compression ratio for different layers of a network. 
On a different context, more recently, \cite{ahn2020chameleon} leverages reinforcement learning for compiler optimization of deep learning models.

Here, we employ RL in the context of quantization to choose an appropriate quantization bitwidth for each layer of a network.

\vspace{0.1cm}
\niparagraph{Training algorithms for quantized neural networks.}
There have been several techniques~\cite{Zhou2016DoReFaNetTL, Zhu2016TrainedTQ, Mishra2017WRPNWR} that train a neural network in a quantized domain after the bitwidth of the layers is determined manually.
DoReFa-Net~\cite{Zhou2016DoReFaNetTL} trains quantized convolutional neural networks with parameter gradients which are stochastically quantized to low bitwidth numbers before they are propagated to the convolution layers. \cite{Mishra2017WRPNWR} introduces a scheme to train networks from scratch using reduced-precision activations by decreasing the precision of both activations and weights and increasing the number of filter maps in a layer. 
DCQ \cite{elthakeb2019divide} proposes an unorthodox method to train quantized neural networks. The proposed approach utilizes knowledge distillation through teacher-student paradigm in a novel setting that exploits the feature extraction capability of DNNs for higher-accuracy quantization. This divide and conquer strategy makes the training of each student section possible in isolation while all these independently trained sections are later stitched together to form the equivalent fully quantized network.
\cite{Zhu2016TrainedTQ} performs the training phase of the network in full precision, but for inference uses ternary weight assignments. For this assignment, the weights are quantized using two scaling factors which are learned during training phase.
PACT \cite{Choi2018PACTPC}  introduces a quantization scheme for activations, where the variable $\alpha$ is the clipping level and is determined through a gradient descent based method.
SinReQ~\cite{elthakeb2019sinreq} proposes a novel sinusoidal regularization for deep quantized training. The proposed regularization is realized by adding a periodic function (sinusoidal regularizer) to the original objective function. By exploiting the inherent periodicity and local convexity profile in sinusoidal functions, SinReQ automatically propel weights towards target quantization levels during conventional training.
Leveraging the sinusoidal properties further, \cite{elthakeb2020gradientbased} extended SinReQ to learn the quantization bitwidth during gradient-based training process. The key insight is that they leverage the observation that sinusoidal period is a continuous valued parameter. As such, the sinusoidal period serves as an ideal optimization objective and a proxy to minimize the actual quantization bitwidth, which avoids the issues of gradient-based optimization for discrete valued parameters.

ReLeQ is an orthogonal technique with a different objective: automatically finding the level of quantization that preserves accuracy and can potentially use any of these training algorithms.

\vspace{0.1cm}
\niparagraph{Ternary and binary neural networks.}
Extensive work, \cite{Hubara2017QNN, Rastegari2016XNORNetIC, Li2016TernaryWN} focuses on binarized neural networks, which impose accuracy loss but reduce the bitwidth to lowest possible level.
In BinaryNet \cite{NIPS2016BNN}, an extreme case, a method is proposed for training binarized neural networks which reduce memory size, accesses and computation intensity at the cost of accuracy.
XNOR-Net \cite{Rastegari2016XNORNetIC} leverages binary operations (such as XNOR) to approximate convolution in binarized neural networks. 
%
%
Another work \cite{Li2016TernaryWN} introduces ternary-weight networks, in which the weights are quantized to {-1, 0, +1} values by minimizing the Euclidian distance between full-precision weights and their ternary assigned values.
However, most of these methods rely on handcrafted optimization techniques and ad-hoc manipulation of the underlying network architecture that are not easily extendable for new networks. 
For example, multiplying the outputs with a scale factor to recover the dynamic range (i.e., the weights effectively become -w and w, where w is the average of the absolute values of the
weights in the filter), keeping the first and last layers at 32-bit floating point precision, and performing normalization before convolution to reduce the dynamic range of the activations.
Moreover, these methods~\cite{Rastegari2016XNORNetIC, Li2016TernaryWN} are customized for a single bitwidth, binary only or ternary only in the case of~\cite{Rastegari2016XNORNetIC} or~\cite{Li2016TernaryWN}, respectively, which imposes a blunt constraint on inherently different layers with different requirements resulting in sub-optimal quantization solutions.
\releq aims to utilize the levels between binary and 8 bits to avoid loss of accuracy while offering automation.
%


\vspace{0.1cm}
\niparagraph{Techniques for selecting quantization levels.}
Recent work ADMM~\cite{Ye2018AUF} runs a binary search to minimize the total square quantization error in order to decide the quantization levels for the layers. Then, they use an iterative optimization technique for fine-tuning.
%
NVIDIA also released an automatic mixed precision (AMP)~\cite{nvidia_auto} which employs mixed precision during training by automatically selecting between two floating point (FP) representations (FP16 or FP32).
%
%
There is a concurrent work HAQ~\cite{HAQ} which also uses RL in the context of quantization.
The following highlights some of the differences.
\releq uses a unique reward formulation and shaping that enables simultaneously optimizing for two objectives (accuracy and reduced computation with lower-bitwidth) within a unified RL process.
In contrast, HAQ utilizes accuracy in the reward formulation and then adjusts the RL solution through an approach that sequentially decreases the layer bitwidths to stay within a predefined resource budget.
This approach also makes HAQ focused more towards a specific hardware platform whereas we are after a strategy than can generalize.
%
%
Additionally, we also provide a systemic study of different design decisions, and have significant performance gain across diverse well known benchmarks.
%
%
The initial version of our work~\cite{ReLeQ}, predates HAQ, and it is the first to use RL for quantization.
Later HAQ was published in CVPR \cite{DBLP:conf/cvpr/WangLLLH19}, and we published initial version of \releq in NeurIPS ML for Systems Workshop \cite{releq_nips19_workshop}.

%% file: body/conclusion.tex

\section{Conclusion}
\vspace{-1ex}
Quantization of neural networks offers significant promise in reducing their compute and storage cost.
However, the utility of quantization hinges upon automating its process while preserving accuracy.
This paper set out to define the automated discovery of quantization levels for the layers while complying to the constraint of maintaining the accuracy. 
As such, this work offered the RL framework that was able to  effectively navigate the huge search space of quantization and automatically quantize a variety of networks leading to significant performance and energy benefits.
The results suggests that a diligent design of our RL framework, which considers multiple concurrent objectives can automatically yield high-accuracy, yet deeply quantized, networks. 

%
%